\documentclass{article}

    \PassOptionsToPackage{numbers, compress}{natbib}
\usepackage[preprint]{edited_neurips_2024}
\usepackage[utf8]{inputenc}        % allow utf-8 input
\usepackage[T1]{fontenc}           % use 8-bit T1 fonts
\usepackage[hidelinks]{hyperref}   % hyperlinks
\usepackage{url}                   % url
\usepackage{graphicx}
\usepackage{booktabs, multicol}    % table making
\usepackage{enumitem}              % adjust margins for enum
\usepackage{wrapfig}               % wrap around table or figure
\usepackage[table, svgnames]{xcolor}  % color table
\usepackage{amssymb, amsmath, amsfonts, amsthm}
\usepackage[toc,page]{appendix}
\usepackage{titletoc}              % table of contents in appendix
\usepackage{algorithm}             % algorithm block
\usepackage{algpseudocode}         % algorithm block
\usepackage{tikz}                  % fancy symbols
\usepackage[labeled,resetlabels]{multibib}

\usepackage{tikz}
\definecolor{YaleBlue}{rgb}{0.059,0.302,0.573}
\definecolor{forestgreen}{rgb}{0.133,0.549,0.133}
\definecolor{crimson}{rgb}{0.863,0.078,0.235}

\newcommand{\tikzxmark}{%
\tikz[scale=0.23] {
    \draw[color=crimson!100, line width=1.4,line cap=round] (0,0) to [bend left=6] (1,1);
    \draw[color=crimson!100, line width=1.4,line cap=round] (0.2,0.95) to [bend right=3] (0.8,0.05);
}}
\newcommand{\tikzcmark}{%
\tikz[scale=0.23] {
    \draw[color=forestgreen!100, line width=1.4,line cap=round] (0.25,0) to [bend left=10] (1,1);
    \draw[color=forestgreen!100, line width=1.6,line cap=round] (0,0.35) to [bend right=1] (0.23,0);
}}

\newcites{L}{Technical Appendices}

\begin{document}

\title{CUTS: A Deep Learning and Topological \\Framework for Multigranular \\Unsupervised Medical Image Segmentation}

\author{
\textbf{Chen Liu}$^{1 *}$ \quad
\textbf{Matthew Amodio}$^{1 *}$ \quad
\textbf{Liangbo L. Shen}$^{2}$ \quad
\textbf{Feng Gao}$^{3}$ \quad
\textbf{Arman Avesta}$^{4}$\\
\textbf{Sanjay Aneja}$^{4 \S}$ \quad
\textbf{Jay C. Wang}$^{5,6 \S}$ \quad
\textbf{Lucian V. Del Priore}$^{5 \S}$ \quad
\textbf{Smita Krishnaswamy}$^{1,3 \S}$ \vspace{6pt}\\
{\small $^1$Yale University Department of Computer Science}\\
{\small $^2$University of California, San Francisco, Department of Ophthalmology}\\
{\small $^{3}$Yale University Department of Genetics \quad
$^{4}$Yale University Department of Therapeutic Radiology}\\
{\small $^{5}$Yale University Department of Ophthalmology \quad
 $^{6}$Northern California Retina Vitreous Associates \vspace{6pt}}\\
{\small *~These authors are joint first authors. \quad $\S$~Senior authors.}\\
{\small Please direct correspondence to: \url{smita.krishnaswamy@yale.edu} or \url{lucian.delpriore@yale.edu}.}
}

% This block is used to define \inserttitle
\makeatletter
\let\inserttitle\@title
\makeatother

% Define colors
\definecolor{wine}{HTML}{830E0D}
\definecolor{color_ode}{RGB}{186, 108, 73}
\definecolor{color_unet}{HTML}{1365C0}
\definecolor{nicered}{HTML}{B22222}
\definecolor{niceblue}{HTML}{0000FF}
% Color table in cell.
\newcommand{\cc}[0]{\cellcolor{color_unet!10}}

% Define a custom command for fancy numbers
\newcommand{\fancynumber}[1]{%
  \tikz[baseline=(char.base)]{
    \node[shape=circle,draw=black,fill=wine,inner sep=1pt](char){\color{white}#1};
  }%
}

% Math related stuff.
\newtheorem{theorem}{Theorem}[section]
\newtheorem{lemma}[theorem]{Lemma}
\newtheorem{prop}[theorem]{Proposition}
\newtheorem{cor}{Corollary}
\newtheorem{definition}{Definition}[section]
\newtheorem{conj}{Conjecture}[section]
\newtheorem{rem}{Remark}

\maketitle
\begin{abstract}
Segmenting medical images is critical to facilitating both patient diagnoses and quantitative research. A major limiting factor is the lack of labeled data, as obtaining expert annotations for each new set of imaging data and task can be labor intensive and inconsistent among annotators. We present CUTS, an unsupervised deep learning framework for medical image segmentation. CUTS operates in two stages. For each image, it produces an embedding map via intra-image contrastive learning and local patch reconstruction. Then, these embeddings are partitioned at dynamic granularity levels that correspond to the data topology. CUTS yields a series of coarse-to-fine-grained segmentations that highlight features at various granularities. We applied CUTS to retinal fundus images and two types of brain MRI images to delineate structures and patterns at different scales. When evaluated against predefined anatomical masks, CUTS improved the dice coefficient and Hausdorff distance by at least 10\% compared to existing unsupervised methods. Finally, CUTS showed performance on par with Segment Anything Models (SAM, MedSAM, SAM-Med2D) pre-trained on gigantic labeled datasets. \\The code is available at \url{https://github.com/KrishnaswamyLab/CUTS}.

\end{abstract}

\section{Introduction}

Medical image segmentation plays an increasingly crucial role in both research and clinical settings in a wide array of imaging modalities including microscopy, X-ray, ultrasound, optical coherence tomography (OCT), computed tomography (CT), magnetic resonance imaging (MRI), positron emission tomography (PET), and others~\cite{fu2021review}. With high-quality image segmentation, clinicians can more easily diagnose and monitor the progression of diseases to improve patient care. Traditional medical image segmentation methods rely on hand-crafted features~\cite{tradMISfeature1,tradMISfeature2,tradMISfeature3, Watershed, Watershed2, Felzenszwalb} or predefined atlases~\cite{tradMISatlas1,tradMISatlas2,tradMISatlas3}. These methods are gradually being replaced by deep learning~\cite{DeepLearning,DLinHealthcare,DLMIS,DLIS} as supervised neural networks demonstrate superior performance than feature-based methods and less overhead than atlas-based methods. Although supervised neural networks have been widely successful in image segmentation in recent years, there are several issues in applying them to medical images, particularly in order to make clinical inferences. First, these networks depend on expert annotations, so they require a large number of labels to adequately cover the data variance to produce reliable segmentations~\cite{DLinHealthcare}. Second, supervised networks trained on one set of annotated images can fail to generalize to similar images collected in very slightly different contexts, such as in different patient populations or on different devices~\cite{GeneralizationProblem}. Third, the desired segmentation granularity may vary across use cases even if the exact same image is concerned --- for example, localizing a brain tumor would require a finer segmentation compared to measuring the brain volume --- yet this need is not easily accommodated by supervised approaches without updating the labels. 
 
 %In practice, the common solution is to repeatedly annotate a sufficiently large subset of images and re-train the networks for each new dataset and/or new segmentation target. However, the annotation process may be cost prohibitive and time-consuming.

To address these issues, we propose to automatically segment medical images using an entirely unsupervised framework that combines recent advances in representation learning with advances in data geometry and topology. An unsupervised approach circumvents the need for costly expert annotations and alleviates the cross-domain generalization problem. More importantly, we also design our approach to produce multigranular segmentations, which can potentially target multiple regions of interest without supervision.

Our framework, which we denote \textbf{C}ontrastive and \textbf{U}nsupervised \textbf{T}raining for multigranular medical image \textbf{S}egmentation (\textbf{CUTS}), was named as an homage to the renowned painter Henri Matisse, who famously used a ``cut-up'' method he called ``drawing with scissors'' to assemble an image based on patches of material from different sources. Our technique is in essence the reverse of this process, as we start with the initial image and use unsupervised machine learning to segment the initial figure into a collection of relatively homogeneous patches using data coarse graining in a learned latent space. Although it may seem trivial to identify the different pieces of paper cut up by scissors, segmentation of medical images is more challenging as the boundaries between biological structures, such as between healthy and pathological tissues, are not always sharp and clean.

CUTS is designed as an unsupervised segmentation pipeline. The images are processed in units of pixel-centered patches, which consists of a fixed-size crop of image centered on an image pixel. A convolutional encoder is then trained on these pixel-centered patches with both intra-image contrastive learning and local patch reconstruction as optimization objectives. We note that contrastive patches should come from the domain of the medical image itself to create a meaningful pixel embedding. Thus, we find suitable contrastive patches within each image itself using an image similarity metric. Subsequently, the learned embedding space serves as a stronger feature-rich foundation for a multiscale, topology-based data coarse graining method called diffusion condensation that produces multigranular segmentations.

\vskip 12pt
Our main contributions include:
\begin{itemize}
    \item CUTS, a novel unsupervised framework with a two-stage approach: it first produces an image-specific pixel-centered patch embedding via a convolutional encoder, and subsequently uses diffusion condensation to coarse-grain these patches into clusters at various levels of granularity to perform multiscale segmentation.
    \item (Specific to the first stage) A novel optimization objective that combines \textbf{intra-image contrastive learning} with local patch reconstruction to help the convolutional encoder learn an expressive embedding space.
    \item (Specific to the second stage) A multiscale cluster assignment approach that utilizes diffusion condensation, which provides clinicians with labels at \textbf{multiple levels of segmentation granularities}, potentially highlighting clinically relevant regions at various scales.
\end{itemize}

The remainder of this paper is organized as follows. First, we discuss related work in the field. With this provided context, we then introduce our framework detailing the neural network architecture, optimization objective, and multiscale segmentation. Finally, we apply our framework on a series of medical image datasets consisting of retinal fundus images and brain MRI images We evaluate the performance of CUTS through qualitative and quantitative metrics and compare to several baselines including other unsupervised approaches and supervised approaches.

\section{Related Works}

\paragraph{Traditional methods for medical image segmentation}
Traditional image segmentation methods generally fall into two categories. The first category relies on hand-crafted image features, such as line/edge detection~\cite{tradMISfeature1}, graph cuts~\cite{tradMISfeature2, Felzenszwalb}, active contours~\cite{tradMISfeature3}, watershed~\cite{Watershed, Watershed2}, level-set~\cite{tradMISfeature4}, and feature clustering~\cite{SLIC}. The methods in this category are simple to execute, but they usually struggle with images with more complicated colors and textures. The second category utilizes a precomputed and annotated atlas to propagate prior knowledge, by warping a predefined set of labels onto new images through image registration~\cite{tradMISatlas1, tradMISatlas2, tradMISatlas3}. These methods require building and annotating an atlas for each image dataset -- a time-consuming process that may sometimes be impractical.

\paragraph{Supervised learning for medical image segmentation}
Supervised deep learning has outperformed traditional methods in segmenting medical images in the past decade~\cite{DLMIS}. In supervised deep learning, a neural network learns to perform a designated task through a data-driven parameter optimization process~\cite{DeepLearning}. In medical image segmentation, the most well-known method is U-Net~\cite{UNet}, followed by a proliferation of variants with skip connections, attention mechanisms, etc.~\cite{zhou2018unet++,oktay2018attention,cciccek20163d,kohl2018probabilistic,yu2023unest,nnUNet}. They are all supervised learning methods and thereby require an abundance of expert annotations.

\paragraph{Towards unsupervised learning}

With a growing emphasis on avoiding reliance on human expert annotations, researchers have been exploring unsupervised learning approaches for medical image segmentation. Many works focused on training with fewer data~\cite{fewshot1, fewshot2, fewshot3, fewshot4, fewshot5}. SSL-ALPNet~\cite{SSL_ALPNet} proposed to directly learn from pseudo-labels generated from Felzenswalb segmentation~\cite{Felzenszwalb}, thus categorizing it as an unsupervised learning method despite a supervised learning approach. DCGN~\cite{DCGN} used a constrained Gaussian mixture model to cluster pixel representations in histopathology images. It assumes that different tissue types correspond to different colors, which is not necessarily true in many other medical image modalities.

Atlas-based unsupervised learning is another promising direction. Compared to their traditional counterparts~\cite{tradMISatlas1, tradMISatlas2, tradMISatlas3}, the versions empowered by deep learning~\cite{AugmentTransformation, LTNet} have improved results. When the domain gap is small, they can be highly effective; otherwise, these methods could fail similarly. Given their requirement for spatial registration, they are more suitable for clearly defined structures that show little variation among individuals and thus are less applicable to image domains with greater variability.

\paragraph{Contrastive learning}
Contrastive learning~\cite{simclr} was proposed as a generic self-supervised method to address the issue of limited annotations. Conceptually, it allows neural networks to learn meaningful representations in the embedding space by encouraging similar image pairs to be embedded closer to each other and vice versa. After a meaningful embedding space is trained, additional layers can be attached and fine-tuned for downstream tasks. In particular, commonly used contrastive learning methods such as SimCLR~\cite{simclr}, SwaV~\cite{swav}, MoCo~\cite{moco}, BYOL~\cite{byol}, BarlowTwins~\cite{barlowtwins} and SimSiam~\cite{simsiam} focus on extracting image-level representations with an inter-image contrastive objective. These image-level contrastive learning methods yield no information about intra-image features and are therefore unsuitable for tasks that require closer scrutiny within the same image, such as image segmentation. In an attempt to adapt contrastive learning to tackle the image segmentation task, \cite{chaitanya2020contrastive} proposed learning image and patch representations through global and local contrastive training. In~\cite{yan2020selfsupervised}, the authors used a similar approach, although they coined different terminologies. Both methods include a supervised fine-tuning stage after contrastive pre-training, which still depends on labels.

\paragraph{Unsupervised image segmentation with contrastive learning}
Two leading unsupervised image segmentation methods, DFC~\cite{DFC} and STEGO~\cite{STEGO}, both utilize contrastive learning concepts. STEGO learns feature relationships between an image and itself, its $k$ most similar images, and dissimilar images. Although STEGO can be trained without labels, it relies on pre-trained vision backbones for knowledge distillation, which is not a requirement in our method. DFC is by far the most similar to our approach, yet with two key differences. First, DFC contrasts on pixels, while we operate on pixel-centered patches. Pixel-centered patches contain significantly richer semantic and textural information than pixels. Second, we achieve segmentation through a topological multiscale coarse-graining method that produces many segmentation maps at various granularities rather than a single segmentation map.

\paragraph{Segment Anything Model~(SAM) and medical variants}
Segment Anything Model~(SAM)~\cite{SAM_Meta} recently introduced a general-purpose segmentation tool pre-trained on a gigantic dataset of natural images. As previous researchers have shown~\cite{SAM_MedIA}, SAM offers an alternative solution to label-free medical image segmentation through an interface called ``zero-shot transfer'', where a single point is provided as a prompt which is deciphered by a prompt encoder and sent to a mask model to produce a segmentation mask. Alternative input formats, such as text prompt (written text) or box prompt (bounding box) are also supported by this framework.

To better adapt to medical image applications, researchers have developed counterparts that are pre-trained on large datasets of medical images instead of natural images. MedSAM~\cite{MedSAM} and SAM-Med2D~\cite{SAM_Med2D} are among the most popular variants.

Strictly speaking, SAM and its variants are not unsupervised learning methods. As a result, they still face the cross-domain generalization problem as previously mentioned, while their brute-force solution is to cover the entire data distribution with the huge training set. Despite their non-unsupervised nature, we decided to include them for comparison, since they are arguably the latest state-of-the-art segmentation framework.

\section{Methods}

\begin{figure*}[!thb]
    \centering
    \includegraphics[width=\textwidth]{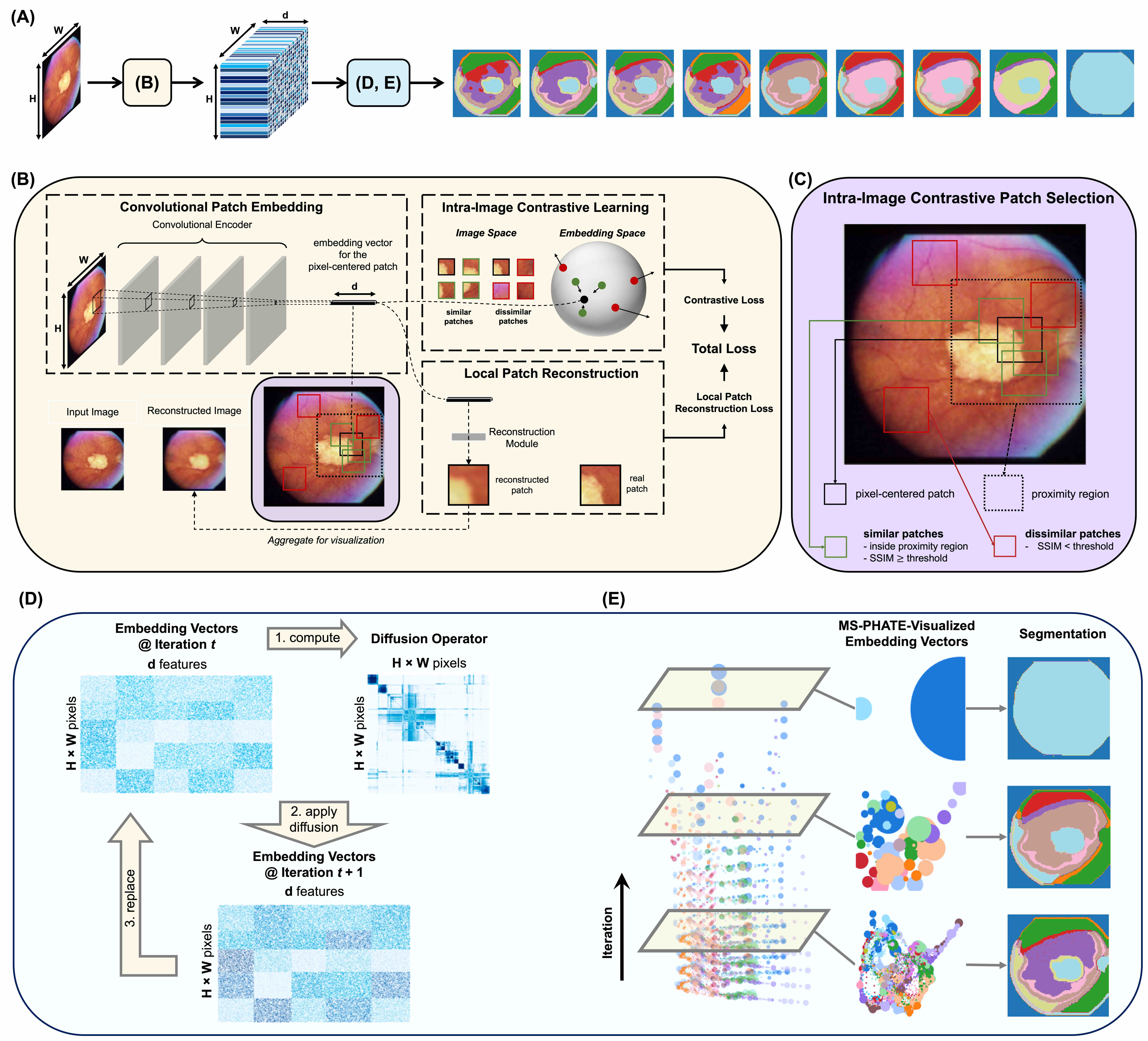}
    \caption{The CUTS Framework. \textbf{(A)} Overview. \textbf{(B)} Pixel-centered patches are mapped into the embedding space, jointly optimized by two objectives. \textbf{(C)} Positive and negative patch pairs are selected based on proximity and structural similarity. \textbf{(D)} Diffusion condensation coarse grains embedding vectors at a series of granularities. \textbf{(E)} Segmentation for any granularity can be performed by mapping cluster assignments to the image space. Multiscale PHATE~(MS-PHATE)~\cite{MS_PHATE} is used for visualization.}
    \label{fig:architecture}
\end{figure*}

The CUTS framework contains two stages~(Fig.~\ref{fig:architecture}(A)). In the first stage, it encodes each pixel along with the local neighborhood around it, denoted as a ``pixel-centered patch'', into a high-dimensional embedding space by jointly optimizing contrastive learning and autoencoding objectives~(Fig.~\ref{fig:architecture}(B)). Unlike most contrastive learning methods that learn from augmented versions of full images, CUTS learns from regions within the same image~(Fig.~\ref{fig:architecture}(C)). This emphasizes learning of local, intra-image features instead of invariance over known image transformations or noise models. This is especially critical for medical images, since they are globally homogeneous (i.e., images from different participants capture the same body part) yet locally heterogeneous (i.e., nuances in structures or textures within small areas of the image are essential). In the second stage, these embedding vectors are coarse-grained to many levels of granularity by diffusion condensation~\cite{brugnone2019coarse, CATCH}. Metastable granularities can be automatically identified from the condensation homology as granularities with zero topological activity~\cite{CATCH}. Segmentation is performed by assigning labels to pixels that correspond to clusters arising from a particular metastable granularity~(Fig.~\ref{fig:architecture}(D-E)).

\subsection{Learning an embedding space for pixel-centered patches}
CUTS uses a convolutional neural network as a patch encoder to map pixel-centered patches from the image space to a latent embedding space. It has convolution, batch norm, activation but no pooling -- to ensure identical spatial dimension between the image and feature map. Two objectives are jointly optimized.

\paragraph{Intra-image contrastive loss}
\label{sec:contrastive_loss}
For any anchor patch $\mathcal{P}_{ij} \in \mathbb{R}^{p \times p \times c}$ centered at coordinates $(i, j)$, we sample positive patches $\{\mathcal{P}^+_{ij}\}$ and negative patches $\{\mathcal{P}^-_{ij}\}$. Let $f$ denote the convolutional encoder. Anchor embedding $z_{ij} = f(\mathcal{P}_{ij})$, positive embeddings $\Omega^+ := \{z^+_{ij}\} = \{f(\mathcal{P}^+_{ij})\}$, and negative embeddings $\Omega^- := \{z^-_{ij}\} = \{f(\mathcal{P}^-_{ij})\}$. After projecting the patches to the latent embedding space, we can perform contrastive learning on their respective embedding vectors $z^+_{ij}$ and $z^-_{ij}$. We mine these positive and negative patches using a combination of a proximity heuristic and an image similarity metric. Only patches nearby (within $\pm$ one patch size) and structurally similar (SSIM~\cite{hore2010image} $> 0.5$) to the anchor patch are considered positive patches. The contrastive loss is defined by Eqn~\eqref{eqn:contrastive} where $sim(\cdot)$ denotes cosine similarity. 

\begin{equation}
    l_{contrast} =
    -\log \frac{\textrm{pos}}{\textrm{neg}}, \quad \textrm{pos} = \sum_{z^+_{ij}\in{\Omega}^+} e^{sim(z_{ij},z^+_{ij})/\tau}, \quad \textrm{neg} = \sum_{z^-_{ij}\in{\Omega}^-} e^{sim(z_{ij},z^-_{ij})/\tau}
    \label{eqn:contrastive}
\end{equation}

\paragraph{Local patch reconstruction loss}
In addition to the contrastive loss, we ensure that our embedding of each pixel-centered patch retains information about the patch around it through a reconstruction loss. For an embedding $z_{ij} \in \mathbb{R}^{d}$, the patch reconstruction loss is $l_{recon} = ||\mathcal{P}_{ij} - f_{recon}(z_{ij})||_2^2$, where $f_{recon}(\cdot) : \mathbb{R}^d \rightarrow \mathbb{R}^{p \times p \times c}$ is a patch reconstruction module. In implementation, $f_{recon}(\cdot)$ is a two-layered fully-connected network with ReLU activation.

\paragraph{Final objective function}
The final objective function is a weighted sum of the contrastive loss and reconstruction loss, balanced by a weighting coefficient $\lambda \in [0, 1]$.

\begin{equation}
    loss = \lambda \cdot l_{contrast} + (1 - \lambda) \cdot l_{recon}
\end{equation}

\begin{figure*}[!htb]
\centering
\includegraphics[width=\textwidth]{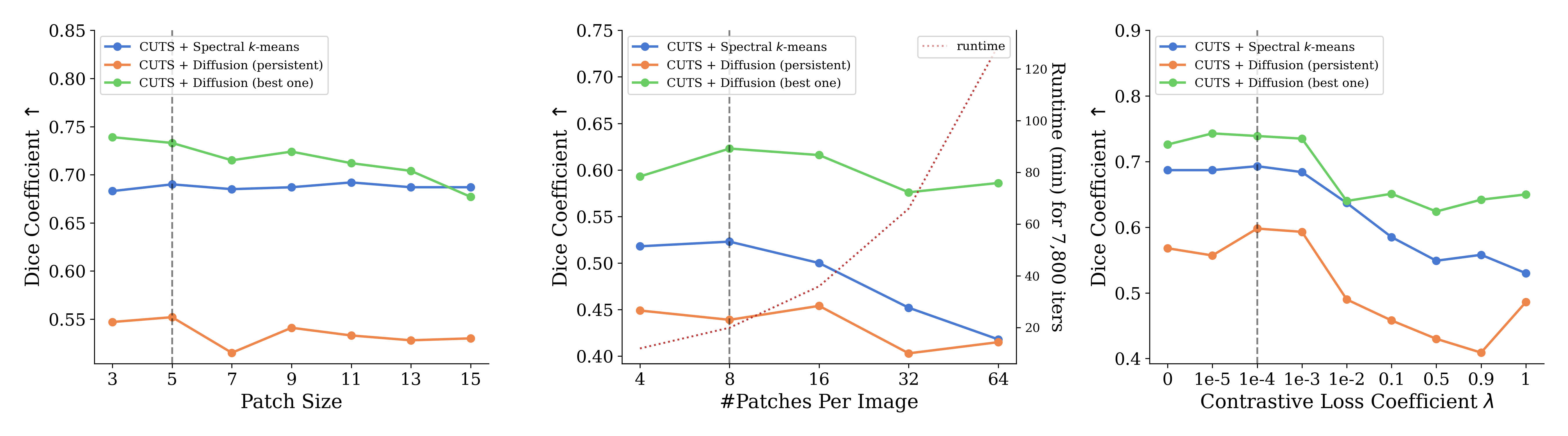}
\caption{Effects of hyperparameters.}
\label{fig:param_tuning}
\end{figure*}

\paragraph{Hyperparameters}
Three key hyperparameters were tuned empirically~(Fig.~\ref{fig:param_tuning}). First, we found that the optimal patch size for pixel-centered patches is $5 \times 5$. Then, we determined to sample $8$ patches in each image for contrastive learning and reconstruction. Lastly, we set the contrastive loss coefficient at $0.0001$. Note that $l_{contrast}$ is still nontrivial after weighing, because the numerical value of $l_{contrast}$ is more than 3 orders of magnitude higher than $l_{recon}$ at convergence.

\subsection{Coarse-graining for multiscale segmentation}
For each image patch $\mathcal{P}_{ij}$ centered at coordinates $(i, j)$, the patch encoder encodes it to $z_{ij} \in \mathbb{R}^d$. We can assign them to $n$ different clusters $\{c_1, c_2, ..., c_n\}$ using a clustering algorithm $cls(\cdot): \mathbb{R}^d \rightarrow \mathbb{R}$. Then, we can create a label map $L \in \mathbb{R}^{H \times W}$ where $L_{ij} = cls(z_{ij})$. The label map $L$ will be the end product of CUTS segmentation. Notably, with diffusion condensation, $cls(\cdot)$ changes throughout the process, and therefore we can generate a rich set of labels.

Diffusion condensation~\cite{brugnone2019coarse, CATCH} is a dynamic process that sweeps through various levels of granularities to identify natural groupings of data. It iteratively condenses data points towards their neighbors through a diffusion process, at a rate defined by the diffusion probability between the points. Unlike most clustering methods, diffusion condensation constructs a full hierarchy of coarse-to-fine granularities where the number of clusters at each granularity is not arbitrarily set but rather inferred from the underlying structure of the data.

Formally, from a data matrix $X^{N \times d}$ with $N$ observations (in our case, $N = W \times H =$ the number of pixels in an image) and $d$ features, we can construct the local affinity among each observation pair $(m, n) \in \{1, ..., N\}$ using a Gaussian kernel

\begin{equation}
    \mathbf{K}(x_m, x_n) = e^{-\frac{||x_m - x_n||^2}{\epsilon}}
    \label{eqn:diffusion_K}
\end{equation}

$\mathbf{K}$ is a $N \times N$ Gram matrix whose $(m, n)$ entry is denoted $\mathbf{K}(x_m, x_n)$ to emphasize the dependency on the data matrix $X$. $x_m$ and $x_n$ are both of dimension $\mathbb{R}^d$. The bandwidth parameter $\epsilon$ controls neighborhood sizes.

Given this affinity matrix $\mathbf{K}$, the diffusion operator is defined by Eqn~\eqref{eqn:diffusion_P} where $\mathbf{D}$ is the diagonal degree matrix as shown in Eqn~\eqref{eqn:diffusion_D}.

\begin{subequations}
\noindent\begin{minipage}[t]{0.4\textwidth}
\begin{equation}
\mathbf{P} = \mathbf{D}^{-1} \mathbf{K}
\label{eqn:diffusion_P}
\end{equation} 
\end{minipage}
\begin{minipage}[t]{0.6\textwidth}
\begin{equation}
\mathbf{D}(x_m, x_m) = \sum_{n} \mathbf{K}(x_m, x_n)
\label{eqn:diffusion_D}
\end{equation}
\end{minipage}
\end{subequations}

The diffusion operator $\mathbf{P}$ defines the single-step transition probabilities for a diffusion process over the data, which can be viewed as a Markovian random walk. To perform multi-step diffusion, one way is to simulate a time-homogeneous diffusion process by raising the diffusion operator to a power of $t$ which leads to $X_t = \mathbf{P}^t X$~\cite{coifman2006diffusion}. On the other hand, as shown in~\cite{CATCH}, we could simulate a time-inhomogeneous diffusion process by iteratively computing the diffusion operator and the data matrix in the following manner.

\begin{align}
\nonumber
&X_0 \leftarrow X \\\nonumber
&\mathbf{for} \hspace{4pt} t \in [1, ..., T]:\\\nonumber
&\hspace{20pt} \mathbf{K}_{t-1} \leftarrow \mathcal{K}(X_{t-1}) \hspace{64pt} \textrm{/* using Eq.~\eqref{eqn:diffusion_K} */}\\
&\hspace{20pt} \mathbf{D}_{t-1} \leftarrow \mathcal{D}(\mathbf{K}_{t-1}) \hspace{63pt} \textrm{/* using Eq.~\eqref{eqn:diffusion_D} */}\\\nonumber
&\hspace{20pt} \mathbf{P}_{t-1} \leftarrow \mathbf{D}_{t-1}^{-1} \mathbf{K}_{t-1}  \hspace{57pt} \textrm{/* using Eq.~\eqref{eqn:diffusion_P} */}\\\nonumber
&\hspace{20pt} X_t \leftarrow \mathbf{P}_{t-1} X_{t-1}
\label{eqn:diffusion_condensation}
\end{align}

The process of diffusion condensation can be summarized as the alternation between the following two steps:
\begin{enumerate}
    \item Computing a time-inhomogeneous diffusion operator from the data at iteration $t$.
    \item Applying this operator to the data, moving points towards the local center of gravity, which forms the data in iteration $t+1$.
\end{enumerate}

More details on diffusion condensation can be found in~\cite{CATCH}. In this paper, we used the official implementation (\url{https://github.com/KrishnaswamyLab/catch}).

We can identify the segments that occur consistently over the series of segmentations, called persistent structures. The terminology ``persistence'' is a measure defined in diffusion condensation as clusters that stay separated over many iterations. The discovery of persistent structures can be achieved by rank-ordering different segments based on their persistence levels, which is quantified by the number of consecutive diffusion iterations in which the segment stays intact and refrains from being merged into another segment.

For binary segmentation, we need to convert the multi-class label maps to binary segmentation masks. Following standard practices~\cite{STEGO, SAM_MedIA}, we use the ground truth segmentation mask to provide a hint on how to select the foreground for each image. Specifically, we iterate over each foreground pixel in the ground truth mask and find the most frequently associated cluster of the corresponding embedding vector. Then we set all pixels whose embeddings match that cluster label as the foreground. This process effectively finds the most probable cluster label if a pixel is randomly selected from the foreground region of the ground truth and thus is objective and unbiased.

\section{Empirical Results}

We prepared three medical image datasets to evaluate our proposed framework. The datasets are chosen to demonstrate the breadth of applications, as they cover variation in color channels (e.g., RGB versus intensity-only), imaging sequences (e.g., T1 versus T2 FLAIR), and organs of interest (e.g., eye versus brain).

\paragraph{Retinal fundus images} We used retinal color fundus images of eyes with Geographic Atrophy (GA) in the age-related eye disease study group~\cite{davis2005age, shen2021relationship}. GA regions were segmented by two graders and reviewed by a retinal specialist, resulting in 56 retinal images with accurate segmentations.

\paragraph{Brain MRI images (ventricles)}
We used MRIs of patients from the Alzheimer's Disease Neuroimaging Initiative study~\cite{LONI}. A radiologist manually segmented the brain ventricles on 100 T1-weighted brain MRIs for our study.

\paragraph{Brain MRI images (tumor)}
We used MRIs of patients with glioma that were scanned by several healthcare facilities. Tumor regions of 200 fluid-attenuated inversion recovery (FLAIR) brain MRIs are segmented by trained medical students and finalized by a board-certified attending neuroradiologist.

\begin{figure*}[!tbh]
    \centering
    \includegraphics[width=\textwidth]{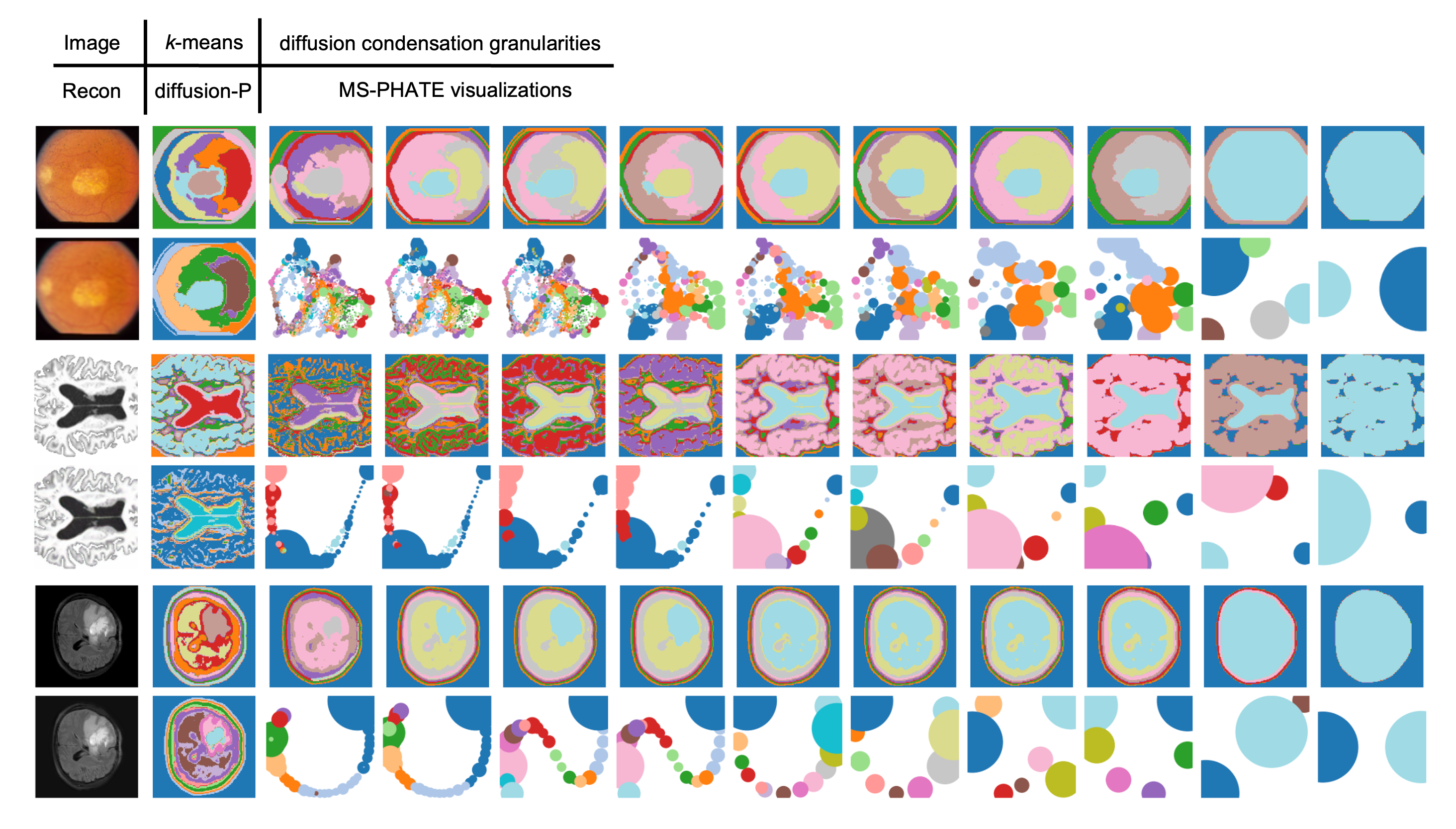}
    \caption{Multigranular segmentation (odd rows) captures distinctive patterns at various scales. Multiscale PHATE (even rows) is used to visualize the diffusion condensation process. The results of CUTS + spectral $k$-means clustering~(``$k$-means'') and CUTS + diffusion condensation persistent structures~(``diffusion-P'') are also shown for reference.}
    \label{fig:results_diffusion}
\end{figure*}

\subsection{Qualitative results on multigranular segmentation}

As shown in Fig.~\ref{fig:results_diffusion}, our multiscale segmentation method provides delineation of image structures at various granularities. The diffusion condensation process starts when all pixels are isolated from each other (pure noise, not shown in the figure). After a few iterations, fine-grained structures begin to emerge, as the most similar pixels are clustered together (leftmost columns starting from the third column). On these finest scales, even the smallest structures are delineated, such as the retinal vessels in the retinal images (first row). Moving toward the coarser scales, anatomical structures arise as tiny patterns collectively form larger groups. Signature structures include the optic disc and geographic atrophy in the retinal images (first row), white and gray matter in the brain ventricles images (third row), and tumor region in the brain tumor images (fifth row). Detection of these anatomical structures can facilitate automatic measurements of their sizes, shapes, and locations for clinical interventions. On the coarser side of the spectrum, most structures are iteratively merged through diffusion condensation, leaving only the most distinctive objects in the image. The final resolution (rightmost column) identifies the two remaining clusters which correspond to the foreground and background, respectively.

Qualitatively, we show that CUTS is able to automatically detect meaningful structures and patterns at multiple granularities within medical images of various modalities. It enables users to determine their desired level of detail without the necessity of manually annotating data for the model's training.

\subsection{Qualitative and quantitative results on binary segmentation}

We compared the performance of CUTS on the three datasets with several alternative methods. We first compared it with three traditional unsupervised methods: Otsu's watershed~\cite{Watershed}, Felzenszwalb~\cite{Felzenszwalb}, and SLIC~\cite{SLIC}. We then compared with DFC~\cite{DFC} and STEGO~\cite{STEGO}, two recent unsupervised models based on deep learning. For each experiment, we re-trained DFC, STEGO, and CUTS on the images only.

Next, we compared against the Segment Anything Model (SAM)~\cite{SAM_Meta, SAM_MedIA} which was pre-trained on 11 million images and 1.1 billion masks, as well as its medical-image variants (MedSAM~\cite{MedSAM}/SAM-Med2D~\cite{SAM_Med2D}) pre-trained on 1.6/4.6 million images and 1.6/19.7 million masks, respectively. For SAM variants, we provided a center point of the ground truth label as a prompt for segmentation of each image~\cite{SAM_MedIA}.

Lastly, for reference, we benchmarked a random labeler as the performance lower bound and two fully supervised methods, UNet~\cite{UNet} and nn-UNet~\cite{nnUNet}, as the upper bound. For coarse-graining of the pixel embeddings, we also implemented a spectral $k$-means clustering~\cite{zha2001spectral} alternative, which segments at only one granularity level. For a fair comparison, we applied the same binarization approach described in the Methods section to \textit{all unsupervised methods}.

\begin{figure*}[!thb]
    \centering
    \includegraphics[width=\textwidth]{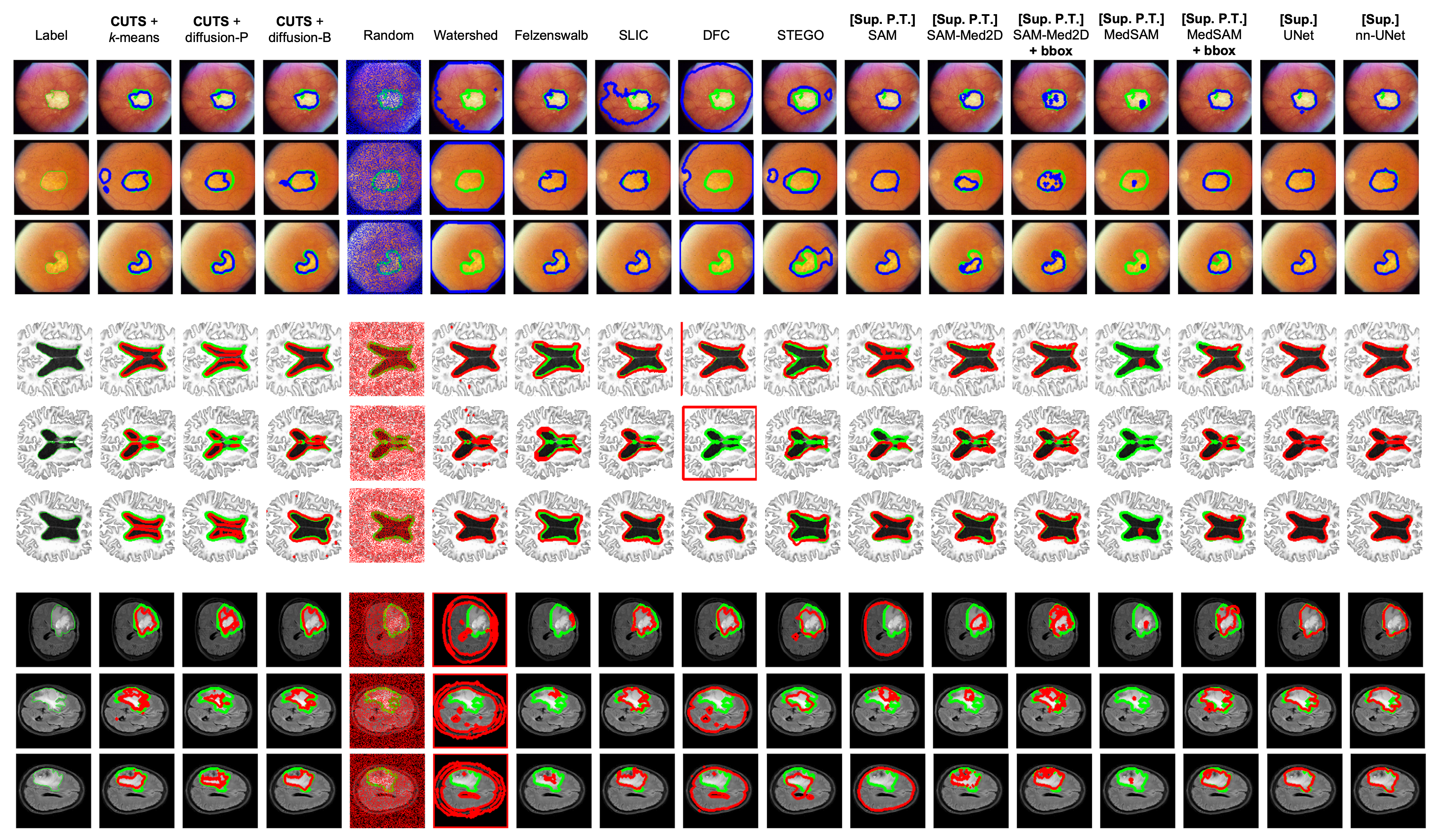}
    \caption{Qualitative segmentation comparison. {\color{ForestGreen}Green} curves outline the ground truth labels while {\color{blue}blue} or {\color{red}red} curves outline the predictions. ``diffusion-B'': the best diffusion condensation granularity. ``\textbf{Sup.}'': supervised ``\textbf{P.T.}'': pre-training. ``\textbf{+bbox}'': using bounding boxes instead of points as input; included for completeness but would be unfair for comparison.}
    \label{fig:results_comparison}
\end{figure*}

\paragraph{Geographic atrophy segmentation in retinal fundus images}

Our first experiment aims to segment regions of geographic atrophy (GA) in retinal fundus images. GA is an advanced stage of age-related macular degeneration (AMD) characterized by progressive macula degeneration. CUTS accurately selects the region of atrophy. 

Qualitatively, CUTS is better at delineating the boundaries of atrophy compared to all other unsupervised methods (Fig.~\ref{fig:results_comparison}). The quantitative results~(Table~\ref{tab:full}) also confirmed this observation. CUTS created better segmentations than other unsupervised methods, as indicated by a higher dice score and a lower Hausdorff distance.

\paragraph{Ventricle segmentation in brain MRI images}

In our next experiment, we tried to segment the brain ventricles in MRI images of patients at various stages of Alzheimer's disease. This task is considered clinically important because the volume of the brain ventricles can predict the progression of dementia~\cite{ventricular_volume_dementia1,ventricular_volume_dementia2}.

Qualitatively, CUTS delineated the brain ventricles in a wide variety of settings~(Fig.~\ref{fig:results_comparison}). Due to the general trend that ventricles appear consistently darker than the rest of the image, most methods are able to achieve good overall performance on several cases. However, our method usually delineates the boundaries better than competing methods, especially for images showing noncontiguous ventricles. The quantitative results~(Table~\ref{tab:full}) also indicate the superior performance of CUTS over other unsupervised methods.

\paragraph{Tumor segmentation in brain MRI images}

Our final experiment investigated a different segmentation target in brain MRI images -- brain tumors, or more specifically, glioma. Accurate segmentation of tumor areas is crucial for the diagnosis and treatment of brain tumors. This process can help radiologists provide vital details about the size, position, and form of tumors, which is important to determine the most appropriate course of clinical care.

Qualitatively, our method demonstrated superior segmentation compared to other unsupervised methods, as shown in Fig.~\ref{fig:results_comparison}. As a general observation, competing methods struggle to identify tumors, although they manage to segment the ventricles in a similar imaging modality. This disparity in performance was anticipated, given the pronounced complexity associated with tumor segmentation compared to ventricles, due to considerably more subtle contrast and morphological distinctions. Nevertheless, CUTS overcomes the inherent challenges and successfully segments tumor regions. CUTS led the other unsupervised methods by a larger margin compared to the less demanding task of ventricle segmentation.

\paragraph{Comparison with SAM, MedSAM and SAM-Med2D} More impressively, as shown in Table~\ref{tab:full}, CUTS achieved better results than every SAM variant on at least 2 out of 3 datasets --- under fair comparison of using a single point as input --- without relying on billions of annotations.

\newcolumntype{x}[1]{>{\centering\arraybackslash\hspace{0pt}}p{#1}}
\renewcommand{\arraystretch}{1.2}

\begin{table*}[!bt]
    \caption{Quantitative comparisons from 3 random seeds. Among unsupervised methods, the best is \textbf{bolded} and runner-up is \underline{underscored}. $^\S$Entries with ``+bbox'' use bounding boxes instead of points as input. They are included for completeness but would be unfair for comparison. $^\ddag$Diffusion condensation will not run since \#features = 1 for each pixel in single-channel images. $^*$The suboptimal performance of MedSAM is expected. According to the authors, ``the point prompt is still an experimental function and the model was trained on a small abdomen CT organ segmentation dataset.''}
    \centering
    \small
    \scalebox{0.57}{
    \begin{tabular}{lx{2.2cm}x{1.8cm}x{2cm}x{2cm}x{2cm}x{2cm}x{2cm}x{2cm}}
    \toprule

    &&
    & \multicolumn{2}{c}{{\large Retinal Atrophy}\hspace{10pt}}
    & \multicolumn{2}{c}{{\large Brain Ventricles}\hspace{10pt}}
    & \multicolumn{2}{c}{{\large Brain Tumor}}\\

    & Deep learning? & Topological? & DSC $\uparrow$ & HD $\downarrow$
    & DSC $\uparrow$ & HD $\downarrow$
    & DSC $\uparrow$ & HD $\downarrow$\\
    \toprule

    \multicolumn{7}{l}{\large \textbf{Unsupervised}, without learning}\\

    Watershed ({\scriptsize IEEE TPAMI$'91$}~\cite{Watershed})
    & $\tikzxmark$ & $\tikzxmark$ 
    & $0.192 {\color{gray}{\pm 0.000}}$
    & $56.32 {\color{gray}{\pm 0.00}}$
    & $\underline{0.781} {\color{gray}{\pm 0.000}}$
    & $30.25 {\color{gray}{\pm 0.00}}$
    & $0.073 {\color{gray}{\pm 0.000}}$
    & $95.42 {\color{gray}{\pm 0.00}}$
    \\

    Felzenszwalb ({\scriptsize IJCV$'04$}~\cite{Felzenszwalb})
    & $\tikzxmark$ & $\tikzxmark$ 
    & $0.592 {\color{gray}{\pm 0.000}}$
    & $27.60 {\color{gray}{\pm 0.00}}$
    & $0.759 {\color{gray}{\pm 0.000}}$
    & $44.80 {\color{gray}{\pm 0.00}}$
    & $0.316 {\color{gray}{\pm 0.000}}$
    & $\textbf{21.41} {\color{gray}{\pm 0.00}}$
    \\

    SLIC ({\scriptsize IEEE TPAMI$'12$}~\cite{SLIC})
    & $\tikzxmark$ & $\tikzxmark$ 
    & $0.567 {\color{gray}{\pm 0.000}}$
    & $28.76 {\color{gray}{\pm 0.00}}$
    & $0.475 {\color{gray}{\pm 0.000}}$
    & $37.96 {\color{gray}{\pm 0.00}}$
    & $0.242 {\color{gray}{\pm 0.000}}$
    & $47.51 {\color{gray}{\pm 0.00}}$
    \\

    \midrule
    \multicolumn{7}{l}{\large \textbf{Unsupervised}, with learning}\\

    DFC ({\scriptsize IEEE TIP$'20$}~\cite{DFC})
    & $\tikzcmark$ & $\tikzxmark$ 
    & $0.300 {\color{gray}{\pm 0.020}}$
    & $46.47 {\color{gray}{\pm 1.42}}$
    & $0.631 {\color{gray}{\pm 0.024}}$
    & $34.28 {\color{gray}{\pm 0.57}}$
    & $0.197 {\color{gray}{\pm 0.004}}$
    & $52.51 {\color{gray}{\pm 0.09}}$
    \\

    STEGO ({\scriptsize ICLR$'22$}~\cite{STEGO})
    & $\tikzcmark$ & $\tikzxmark$ 
    & $0.649 {\color{gray}{\pm 0.025}}$
    & $34.12 {\color{gray}{\pm 4.06}}$
    & $0.725 {\color{gray}{\pm 0.050}}$
    & $12.59 {\color{gray}{\pm 4.43}}$
    & $0.176 {\color{gray}{\pm 0.104}}$
    & $57.16 {\color{gray}{\pm 14.09}}$
    \\

    \rowcolor{YaleBlue!10}
    \textbf{(Ours)} CUTS + Spectral $k$-means
    & $\tikzcmark$ & $\tikzxmark$ 
    & $\underline{0.675} {\color{gray}{\pm 0.014}}$
    & $26.82 {\color{gray}{\pm 0.88}}$
    & $0.774 {\color{gray}{\pm 0.008}}$
    & $\underline{8.31} {\color{gray}{\pm 0.23}}$
    & $\underline{0.432} {\color{gray}{\pm 0.010}}$
    & $33.94 {\color{gray}{\pm 0.65}}$
    \\

    \rowcolor{YaleBlue!10}
    \textbf{(Ours)} CUTS + Diffusion (pers.)
    & $\tikzcmark$ & $\tikzcmark$ 
    & $0.604 {\color{gray}{\pm 0.003}}$
    & $\underline{21.69} {\color{gray}{\pm 0.44}}$
    & $0.495 {\color{gray}{\pm 0.002}}$
    & $13.36 {\color{gray}{\pm 0.60}}$
    & $0.390 {\color{gray}{\pm 0.004}}$
    & $33.66 {\color{gray}{\pm 0.24}}$
    \\

    \rowcolor{YaleBlue!10}
    \textbf{(Ours)} CUTS + Diffusion (best)
    & $\tikzcmark$ & $\tikzcmark$ 
    & $\textbf{0.741} {\color{gray}{\pm 0.007}}$
    & $\textbf{17.76} {\color{gray}{\pm 0.13}}$
    & $\textbf{0.810} {\color{gray}{\pm 0.006}}$
    & $\textbf{7.17} {\color{gray}{\pm 0.18}}$
    & $\textbf{0.486} {\color{gray}{\pm 0.007}}$
    & $\underline{25.16} {\color{gray}{\pm 1.12}}$
    \\

    \midrule
    \multicolumn{7}{l}{\large \textbf{Ablation}: image pixels instead of latent embeddings}\\

    Image pixels + Spectral $k$-means
    & $\tikzxmark$ & $\tikzxmark$ 
    & $0.560 {\color{gray}{\pm 0.000}}$
    & $37.97 {\color{gray}{\pm 0.00}}$
    & $0.386 {\color{gray}{\pm 0.000}}$
    & $26.11 {\color{gray}{\pm 0.00}}$
    & $0.240 {\color{gray}{\pm 0.000}}$
    & $51.69 {\color{gray}{\pm 0.00}}$
    \\

    Image pixels + Diffusion (pers.)
    & $\tikzxmark$ & $\tikzcmark$ 
    & $0.405 {\color{gray}{\pm 0.000}}$
    & $61.67 {\color{gray}{\pm 0.00}}$
    & $\ddag$
    & $\ddag$
    & $\ddag$
    & $\ddag$
    \\

    Image pixels + Diffusion (best)
    & $\tikzxmark$ & $\tikzcmark$ 
    & $0.538 {\color{gray}{\pm 0.000}}$
    & $45.16 {\color{gray}{\pm 0.00}}$
    & $\ddag$
    & $\ddag$
    & $\ddag$
    & $\ddag$
    \\

    \midrule
    \multicolumn{7}{l}{\large \textbf{Lower bound}: random label}\\

    Random
    & $\tikzxmark$ & $\tikzxmark$ 
    & $0.132 {\color{gray}{\pm 0.000}}$
    & $78.45 {\color{gray}{\pm 0.07}}$
    & $0.149 {\color{gray}{\pm 0.000}}$
    & $61.40 {\color{gray}{\pm 0.02}}$
    & $0.057 {\color{gray}{\pm 0.000}}$
    & $95.53 {\color{gray}{\pm 0.02}}$
    \\

    \midrule
    \multicolumn{7}{l}{\large \textbf{Upper bound}: supervised}\\

    SAM ({\scriptsize ICCV$'23$~\cite{SAM_Meta}, MedIA$'23$~\cite{SAM_MedIA}})
    & $\tikzcmark$ & $\tikzxmark$ 
    & $0.924 {\color{gray}{\pm 0.000}}$
    & $9.18 {\color{gray}{\pm 0.01}}$
    & $0.644 {\color{gray}{\pm 0.003}}$
    & $30.24 {\color{gray}{\pm 0.19}}$
    & $0.405 {\color{gray}{\pm 0.000}}$
    & $36.14 {\color{gray}{\pm 0.14}}$
    \\

    SAM-Med2D ({\scriptsize ArXiv~\cite{SAM_Med2D}})
    & $\tikzcmark$ & $\tikzxmark$ 
    & $0.548 {\color{gray}{\pm 0.001}}$
    & $14.69 {\color{gray}{\pm 0.00}}$
    & $0.736 {\color{gray}{\pm 0.000}}$
    & $17.38 {\color{gray}{\pm 0.02}}$
    & $0.591 {\color{gray}{\pm 0.001}}$
    & $12.93 {\color{gray}{\pm 0.01}}$
    \\

    % \rowcolor{gray!20}
    {\color{gray!60} SAM-Med2D\textbf{+bbox}}$^\S$
    & $\tikzcmark$ & $\tikzxmark$ 
    & ${\color{gray!60} 0.882 {\pm 0.000}}$
    & ${\color{gray!60} 5.31 {\pm 0.00}}$
    & ${\color{gray!60} 0.849 {\pm 0.000}}$
    & ${\color{gray!60} 9.78 {\pm 0.00}}$
    & ${\color{gray!60} 0.686 {\pm 0.000}}$
    & ${\color{gray!60} 8.74 {\pm 0.00}}$
    \\

    MedSAM$^*$ ({\scriptsize Nat. Commun.$'24$~\cite{MedSAM}})
    & $\tikzcmark$ & $\tikzxmark$ 
    & $0.079 {\color{gray}{\pm 0.000}}$
    & $32.29 {\color{gray}{\pm 0.02}}$
    & $0.053 {\color{gray}{\pm 0.000}}$
    & $64.00 {\color{gray}{\pm 0.04}}$
    & $0.088 {\color{gray}{\pm 0.001}}$
    & $33.54 {\color{gray}{\pm 0.02}}$
    \\

    % \rowcolor{gray!20}
    {\color{gray!60} MedSAM\textbf{+bbox}}$^\S$
    & $\tikzcmark$ & $\tikzxmark$ 
    & ${\color{gray!60} 0.889 {\pm 0.000}}$
    & ${\color{gray!60} 5.21 {\pm 0.00}}$
    & ${\color{gray!60} 0.829 {\pm 0.000}}$
    & ${\color{gray!60} 10.60 {\pm 0.00}}$
    & ${\color{gray!60} 0.702 {\pm 0.000}}$
    & ${\color{gray!60} 7.61 {\pm 0.00}}$
    \\

    UNet ({\scriptsize MICCAI$'15$}~\cite{UNet})
    & $\tikzcmark$ & $\tikzxmark$ 
    & $0.965 {\color{gray}{\pm 0.014}}$
    & $3.78 {\color{gray}{\pm 1.08}}$
    & $0.989 {\color{gray}{\pm 0.001}}$
    & $1.05 {\color{gray}{\pm 0.10}}$
    & $0.867 {\color{gray}{\pm 0.016}}$
    & $8.84 {\color{gray}{\pm 1.10}}$
    \\

    nnUNet ({\scriptsize Nat. Methods$'21$}~\cite{nnUNet})
    & $\tikzcmark$ & $\tikzxmark$ 
    & $0.937 {\color{gray}{\pm 0.014}}$
    & $6.00 {\color{gray}{\pm 1.35}}$
    & $0.984 {\color{gray}{\pm 0.005}}$
    & $2.10 {\color{gray}{\pm 0.42}}$
    & $0.834 {\color{gray}{\pm 0.024}}$
    & $8.64 {\color{gray}{\pm 1.60}}$
    \\
    \bottomrule
    \end{tabular}
}
\label{tab:full}
\end{table*}
\renewcommand{\arraystretch}{1}

\paragraph{Ablation study}
We confirmed that applying diffusion condensation or spectral $k$-means on the raw image pixels is suboptimal compared to CUTS~(Table~\ref{tab:full}).

\section{Conclusion}

CUTS is a deep learning and topological framework that identifies and highlights important medical image structures using unsupervised learning. Despite the emergence of foundation models, such as variants of SAM, CUTS remains relevant and insightful. It is lightweight and does not require extensive annotation and pre-training in large compute warehouses. Additionally, it is clear that foundation models like SAM necessitate domain-specific fine-tuning for tasks not covered by the initial supervised pre-training, which highlights the continued relevance of approaches like CUTS that investigate objectives, modules and techniques to inject the correct inductive biases. Therefore, CUTS offers a practical and effective alternative in the evolving landscape of medical imaging.

\section{Discussion}

Current state-of-the-art methods for medical image segmentation are primarily supervised and therefore require domain experts to annotate a large number of medical images. Moreover, it is often infeasible to collect enough images of rare diseases to train supervised learning models. For example, in this work we studied a retinal degeneration condition. The number of images available to any institution of this condition is usually within a hundred, several orders of magnitude fewer than popular natural image databases for deep learning with millions of images. Furthermore, another limitation of supervised learning approaches is the domain generalization problem. When a method is optimized for a specific type of image used for training, its performance may suffer if used on other types of image, even if they are only slightly dissimilar.

In contrast, unsupervised methods, like CUTS, while more challenging architect,  do not require human expert grading and thereby circumvent this time-consuming, expensive, and labor-intensive initial step. Unsupervised methods can also be applied to much smaller datasets, ideal for rare diseases. Unfortunately, prior attempts to use unsupervised learning to segment medical images have not achieved the desired results. These unsupervised methods often yield subpar performance, despite having advantages including independence from labels and the ability to generalize to new datasets while preserving robustness.

CUTS bridges the difficulty of creating unsupervised images by using the key insight that, while an image as a whole may be hard to segment, pixels forming boundaries of image features may be detectable by their local context. Thus, CUTS features carefully architected losses for local pixel-centered patch reconstruction and pixel-centered patch-based contrastive losses based on within-image augmentation of patches. With these unique penalties, CUTS learns an intermediate representation of a pixel-centered patch embedding for each image. The key advantage of this pixel-centered patch representation is that it is amenable to not one segmentation, but several multigranular segmentations of the same image via a topological coarse-graining scheme. The final output of CUTS is thus several segmentations of the image with features of different resolutions of interest for different clinical queries.

In our brain tumor image dataset, for example, CUTS enables: (1) brain extraction on the coarsest scale, (2) isolation of white matter, gray matter, and cerebrospinal fluid on the intermediate scale, and (3) small tumor segmentation on the finest scale. These different features can be important for different diagnostic purposes such as tumor placement identification for surgical purposes or small tumor size extraction for the analysis of metastases.

In this work, we demonstrate the application of CUTS to three medical image datasets from different medical domains. On the retinal fundus images, the watershed, Felzenszwalb, and DFC focus primarily on the contours of the retina without distinguishing the geographic atrophy regions, where the contrast is more subtle. SLIC and STEGO generally perform better, yet they tend to overestimate the region of interest. CUTS avoids all these caveats and consistently segments geographical atrophy. On brain MRI for ventricle segmentation, the watershed method often ignores the frontal or lateral half of the ventricles. Felzenszwalb, SLIC, and DFC have difficulty determining the segmentation boundary. STEGO tends to include tissues around the ventricles. CUTS is slightly more conservative on the boundary regions but nevertheless outperforms other unsupervised methods. On the brain MRI images for tumor segmentation, the watershed and Felzenszwalb methods merely isolate the entire brain from the background with no attention to detailed structures. SLIC, DFC, and STEGO either ignored the tumor region or merged it with the background. CUTS, on the other hand, is sensitive to tenuous contrast transitions in tumor regions and generates significantly better segmentations.

In conclusion, CUTS allows us to identify and highlight important medical image structures using an unsupervised learning approach. This has enormous implications for the expanding field of medical image evaluation and represents a step forward in identifying important and distinct information relevant to the clinical interpretation of images. Such interpretation of medical images is crucial to disease detection in asymptomatic individuals; examples include mammography for screening for breast cancer, teleophthalmology and automated image analysis for screening diabetes for eye disease, or vulnerable populations for macular degeneration and glaucoma, and screening of high-risk populations such as smokers for early lung cancer.

\section{Acknowledgements}
The authors would like to thank Mengyuan Sun, Aneesha Ahluwalia, Benjamin K. Young, and Michael M. Park for delineating geographic atrophy borders on fundus photographs.

This work was supported in part by the National Science Foundation (NSF~DMS~2327211, NSF Career Grant~2047856) and the National Institute of Health (NIH~1R01GM130847-01A1, NIH~1R01GM135929-01).

\clearpage
\newpage
\bibliographystyle{unsrt}
\bibliography{references}

\end{document}